\documentclass{llncs}
\usepackage{makeidx}  
\usepackage{algorithm}
\usepackage[noend]{algpseudocode}
\usepackage{graphicx}
\usepackage[utf8]{inputenc}
\graphicspath{ {images/} }
\usepackage{booktabs}
\begin{document}
\frontmatter          
\pagestyle{headings}  
\addtocmark{Hamiltonian Mechanics} 

\author{Luis Argerich \and
Matias J. Cano \and
Joaquin Torre Zaffaroni}
\authorrunning{Argerich et al.} 
\institute{Facultad de Ingeniería, Universidad de Buenos Aires, \\
Ciudad Autónoma de Buenos Aires, Argentina. \\
\email{lrargerich@gmail.com}}

\mainmatter              
\title{Hash2Vec: Feature Hashing for Word Embeddings}
\titlerunning{Hash2Vec}  
%
\maketitle              

\begin{abstract}
In this paper we propose the application of feature hashing to create word embeddings for natural language processing. Feature hashing has been used successfully to create document vectors in related tasks like document classification. In this work we show that feature hashing can be applied to obtain word embeddings in linear time with the size of the data. The results show that this algorithm, that does not need training, is able to capture the semantic meaning of words. We compare the results against GloVe showing that they are similar. As far as we know this is the first application of feature hashing to the word embeddings problem and the results indicate this is a scalable technique with practical results for NLP applications.
\end{abstract}
\section{Introduction}
The main objective of word embeddings is to form a vector space of words that makes ``sense''. Usually, this means that semantically similar words are together. One use for these language models is to estimate the probability of an n-gram being correct.  \cite{ReverseDictionaryPaper} suggested that word embeddings can also be used to create reverse-dictionaries, in which one writes the definition of a word and the algorithm suggests a concept that fits the definition. Even more, it is possible to create bilingual reverse dictionaries which can be quite useful in translation tasks. In recent works, several interesting properties of the resulting vector spaces were found \cite{MikolovAnalogReasoningTask}. There are many other applications of word embeddings \cite{Glove} which make the topic a very active area of research in NLP. 

One way to create word embeddings is by using the bag of words (BOW) model where the word co-occurrence matrix is calculated. In this matrix, each row represents a unique word so that the i,j-element is the amount of times word $j$ has co-occurred with word $i$. This matrix can become huge in the order of millions by millions, making its use difficult in any application. 

As a result, modern models (GloVe \cite{Glove}, Word2Vec \cite{MikolovPhrases} learn to represent words with a fixed reduced dimensionality. 

A simple technique for dimensionality reduction is Feature Hashing \cite{Shi}, also known as the \textit{hashing trick}. The idea is to apply a hashing function to each feature of a high dimensional vector to determine a new dimension for the feature in a reduced space. Feature hashing has been used successfully to reduce the dimensionality of the BOW model for texts \cite{Shi}. \cite{WeinbergerSpam} used feature hashing to classify mail as \textit{spam} or \textit{ham}.

To mitigate the effect of collisions in the resulting vectors \cite{WeinbergerSign} proposes the use of a second hash function $\xi$ that determines the sign of a feature. 

It has been shown that Feature Hashing preserves the inner product between vectors and the error can be bounded. This is explained using the Johnson-Lindenstrauss lemma \cite{JohnsonLindenstrauss} \cite{Achlioptas} and showing that feature hashing is a particular case of a J-L projection where the projection matrix has exactly one $+1$ or $-1$ in each row.  

Because of this, if we apply the hashing trick to the word co-occurrence matrix we are able to obtain an embedding where the inner products between the embedded vectors accurately represent the inner products between the original vectors in the co-occurrence matrix. Our experiments confirm that the distortion between using the full vectors and this embedding is minimal. This means that vectors that are close in the original matrix will also be close in the embedded space. 

Interestingly, embeddings can be constructed without the need of computing the word full-size matrix. This makes the algorithm efficient, as memory consumption is reduced  from $O(n^2)$ to $O(n\times k)$, where $n$ is the size of the corpus in words and $k$ is a fixed dimensionality that can be small (in the order of hundreds.) 

\section{Algorithm}


The main formula for the algorithm can be seen in (1). It is a variation of the feature hashing equation shown in \cite{Shi} with the addition of the second hashing function proposed by \cite{WeinbergerSign}, as well as the domain-specific part which is the weight function for each co-occurrence. 

\begin{equation}
\bar{w}_j^{(k)} = \sum_{w^{(c)} \in C_k; h(w^{(c)})=j} \xi(w^{(c)})\sum_{i=1}^{n^{(k)}(w^{(c)})} f_i^{(k,c)}
\end{equation}

 $w(k)$ represents the k-th word, and $\bar{w}(k)$ represents the reduced vector for such word and therefore $\bar{w_j}(k)$ is the $j$-th component for the k-th word vector. $C_k$ represents all the contexts of the $k$-th word and $h$ is a hashing function as \cite{Shi} shows. $\xi$ is the additional hashing function such that $\xi: \mathrm{String} \rightarrow \{-1,1\}$ proposed by \cite{WeinbergerSign}. $n^{(k)}(x)$ represents the amount of times word $x$ has appeared with word $k$, and finally, $f_i(k,c)$ is the aging (or weighting) for the word $c$ according to word $k$ in the $i$-th time they have appeared together.

Words are processed from the text in linear fashion and for each new word an embedding vector is created. A window of size $k$ is defined to determine which words co-occur in the context of another. We iterate through the context applying Feature Hashing to construct the embeddings. A simplified version of the algorithm can be seen in Algorithm 1. 

\begin{algorithm}
Parameters: \textit{n} the embedding size, \textit{k} the context size, $h$ hash function, $\xi$ hash sign-function and $f$ aging function.
\caption{Hash2Vec}\label{euclid}
\begin{algorithmic}[1]
\State words $\gets$ $Dictionary()$
\For{ every word $w$ in text}
    \If{$w\ \notin$ keys(words)}
        words[$w$] $\gets$ Array($n$)
    \EndIf
    \For{ every context word $cw$ with distance $d$ }
        \State weight $\gets$ $f(d)$
        \State sign $\gets$ $\xi(cw)$
        \State words[$w$][$h(cw)$] $\gets$ words[$w$][$h(cw)$] $ +\ \mathrm {sign} \times \mathrm{weight}$
    \EndFor
\EndFor

\end{algorithmic}
\end{algorithm}

In practice we only need to keep track of the $k$ past words and update both the embedding of the current one and the previous ones. This scheme allows the embeddings to be computed in a streamlined way with $O(n)$ complexity. For simplicity we assume that for each word the embedding is updated based on the words within the $k$-sized window. 

The weight function $f$ is a parameter and a deciding factor on the performance of the algorithm. For example, if $f(d) = 1\ \forall d$, then we are simply calculating a reduced version of the co-occurrence matrix. We obtained better results using $f$ similar to a Gaussian distribution, i.e., $f(d) = e^{-\left (\frac{d}{\sigma} \right )^2} $.

One interesting property of Hash2Vec is that it always constructs the same embeddings for the same training corpus, while word2vec depends on the starting seed and GloVe uses stochastic optimization. 

\subsection{Variations on the algorithm}
In order to improve the quality of embeddings, we decided to try some variations on the basic idea and performed various preprocessing tasks before running Hash2Vec. 

While running the model with huge corpora there are words that appear frequently but that don't carry meaning as they only serve a syntactical purpose. To prevent these words from being a source of noise for Hash2Vec, we preprocess the corpus to remove them. We used two criteria to select the words to filter: calculating a certain percentile, avoiding the words above it or using a stoplist of words to be removed. Both methods resulted in an improvement on the similarity tests. 

We applied the algorithm proposed on \cite{MikolovPhrases} to adjoin phrases. This is very important because otherwise ``New York'', ``San Francisco'', etc. would not exist as a single token. 

We also obtained modestly better results using \textit{homogeneous sentence selection}, in which each sentence in an original text moves to a final text according to a uniform probability distribution, instead of simply truncating the text. Our hypothesis is that with better sampling, the context words are less biased to local articles (if using a source like Wikipedia) inherent to the corpus, making the final vector better in the sense that it was trained with many different articles and usages of the word. Moreover, our experiments revealed that this algorithm is sensible to polysemy. \cite{SocherPolysemy} explore this idea to construct different representations for the same word.  

\subsection{Applications to speed up other embedding algorithms}

The algorithm that we propose constructs a lower-dimensional approximation of the word co-occurrence matrix. The embeddings can then be used directly or can be the starting point for a matrix factorization algorithm, like the one in \textit{GloVe}. In the case of \textit{word2vec}, \cite{LiFactorization} have shown that the SGNS model learns an explicit matrix factorization (EMF) and that very similar results can be obtained by using applying the EMF on the original matrix. The reduced-form matrix that Hash2Vec creates can be used instead of the original to obtain again very similar results without the memory footprint of an $O(n^2)$ matrix. 

\subsection{Streaming applications and parallelization}

Since the embeddings are refined as text is processed, the algorithm is practical for streaming applications. The model can also adapt to changes in language, e.g. new words being used. Thus the embeddings can constantly be updated while being always usable, which is something GloVe cannot do.

By nature Hash2Vec is easily parallelizable since the embeddings can be updated from different portions of text in parallel and then just added up to construct the final vectors. This associative property makes it trivial for the algorithm to be ported to a model like MapReduce. The algorithm could be run on Apache Spark, where each node processes a fragment of text and a single final reduce operation can output the final embeddings. Even more, if no regularization function is applied on the final vectors, embeddings trained in different contexts can be combined by adding the individual vectors of each word in both corpora.  

\section{Results}
For benchmarking we compared our results with two commonly used datasets. The first is wordsim353 \cite{WordsimPaper} and the second is from Amazon's Mechanical Turk, as used in \cite{MturkPaper}. Both datasets hold word pairs with similarity scores for each, representing human assigned similarity judgements. We then calculate the Spearman correlation between our results and the dataset's to obtain a metric on the quality of the embeddings. The similarity measure we used is cosine similarity, like the one used in \cite{Glove}.

\begin{figure}
    \includegraphics[width=6cm,height=4.5cm]{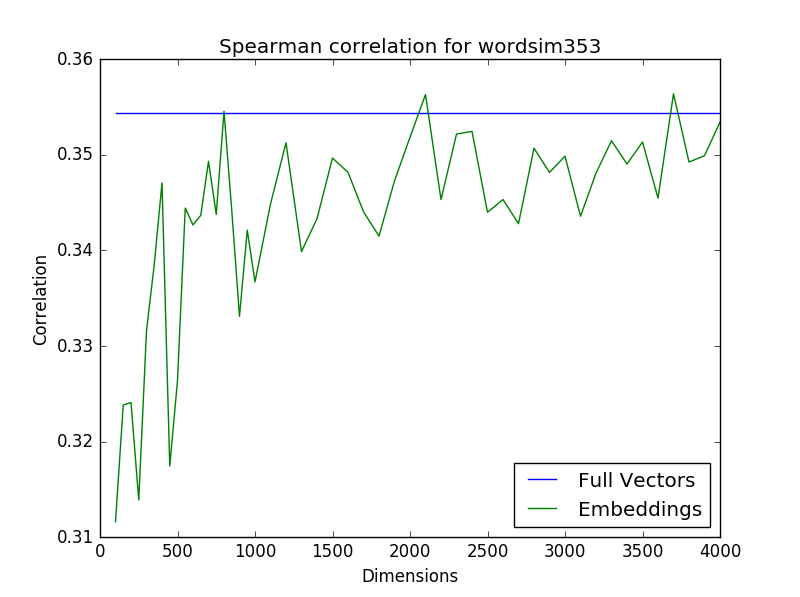}
    \includegraphics[width=6cm,height=4.5cm]{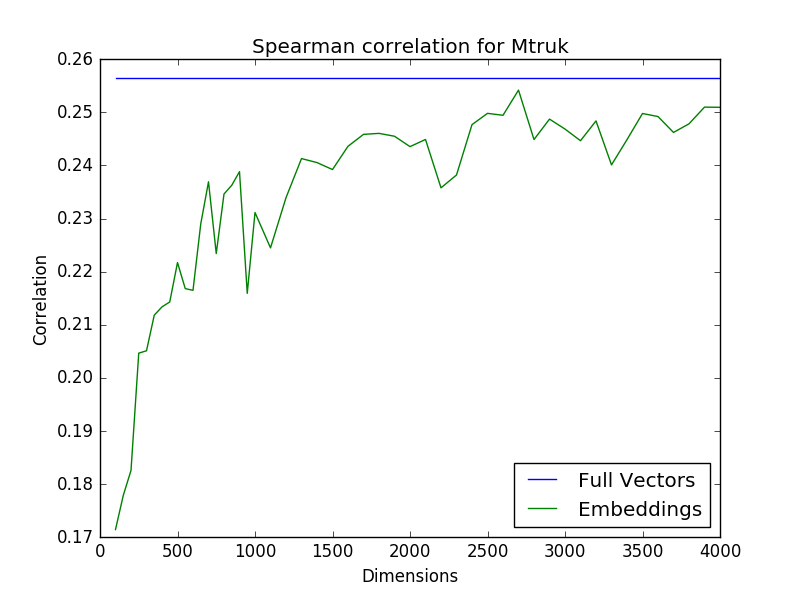}
  \caption{Algorithm's base performance (by correlating it to the Wordsim353 and Amazon Mechanical Turk datasets) as the vector size increases.}
  \label{fig:correlations}
\end{figure}

Both graphs in Figure \ref{fig:correlations} show that Hash2Vec approximates the full vectors when the vector dimension increases. A very good approximation can be obtained with dimensions in the order of the hundreds, orders of magnitude smaller than full vectors. As expected, the co-ocurrence matrix appears to be a theoretical limit on the performance of Hash2Vec.

\subsection{Comparison with GloVe}

We trained GloVe using the same corpus with $k=15$ and $n=600$ in both cases. Table \ref{similarwordstask} shows the 5 most similar words returned by GloVe and Hash2Vec for some queries.

\begin{table}[]
\centering
\caption{Most similar words comparison with GloVe}
\label{similarwordstask}
\begin{tabular}{llllllll}
         & \underline{computer}  & \underline{king}    &  \underline{physics}     & \underline {italy}& \underline {wounded}  & \underline {anglican}     &  \\
\textbf{GloVe}    & computers & son     & chemistry   & germany     & killed   & churches     &  \\
         & software  & i       & mechanics   & france      & mortally & church       &  \\
         & systems   & emperor & quantum     & italian     & soldiers & catholic     &  \\
         & hardware  & ii      & mathematics & greece      & injured  & communion    &  \\
         & game      & kings   & particle    & spain       & dead     & orthodox     &  \\ 
         \\
\textbf{Hash2Vec} & program   & england & mathematics & france      & killed   & lutheran     &  \\
         & computers & iii     & study       & germany     & injured  & episcopal    &  \\
         & hardware  & henry   & chemistry   & spain       & captured & orthodox     &  \\
         & programs  & charles & theory      & switzerland & after    & presbyterian &  \\
         & game      & james   & astronomy   & russia      & defeated & communion    & 
\end{tabular}
\end{table}

We observe the results of GloVe and Hash2Vec to be very similar. Both models capture the semantic meaning of words and in some cases Hash2Vec seems to outperform Glove. 


\subsection{Embeddings visualization}

To understand how the algorithm behaves and to further observe the resulting vector space, we applied t-SNE \cite{tsnepaper}, a widely used dimensionality reduction algorithm. 

After training Hash2Vec on a 110-million-word corpus ($k=5$ and $n=600$), we applied t-SNE. Given that the algorithm groups together similar words, we should be able to see it in the graph. Since t-SNE is about $O(n^2)$ in memory, we only used the $15,000$ most common words. Zipf's law gives some guarantees to only keeping the most common words. 
\begin{figure}
  \centering
    \includegraphics[width=10cm,height=6cm]{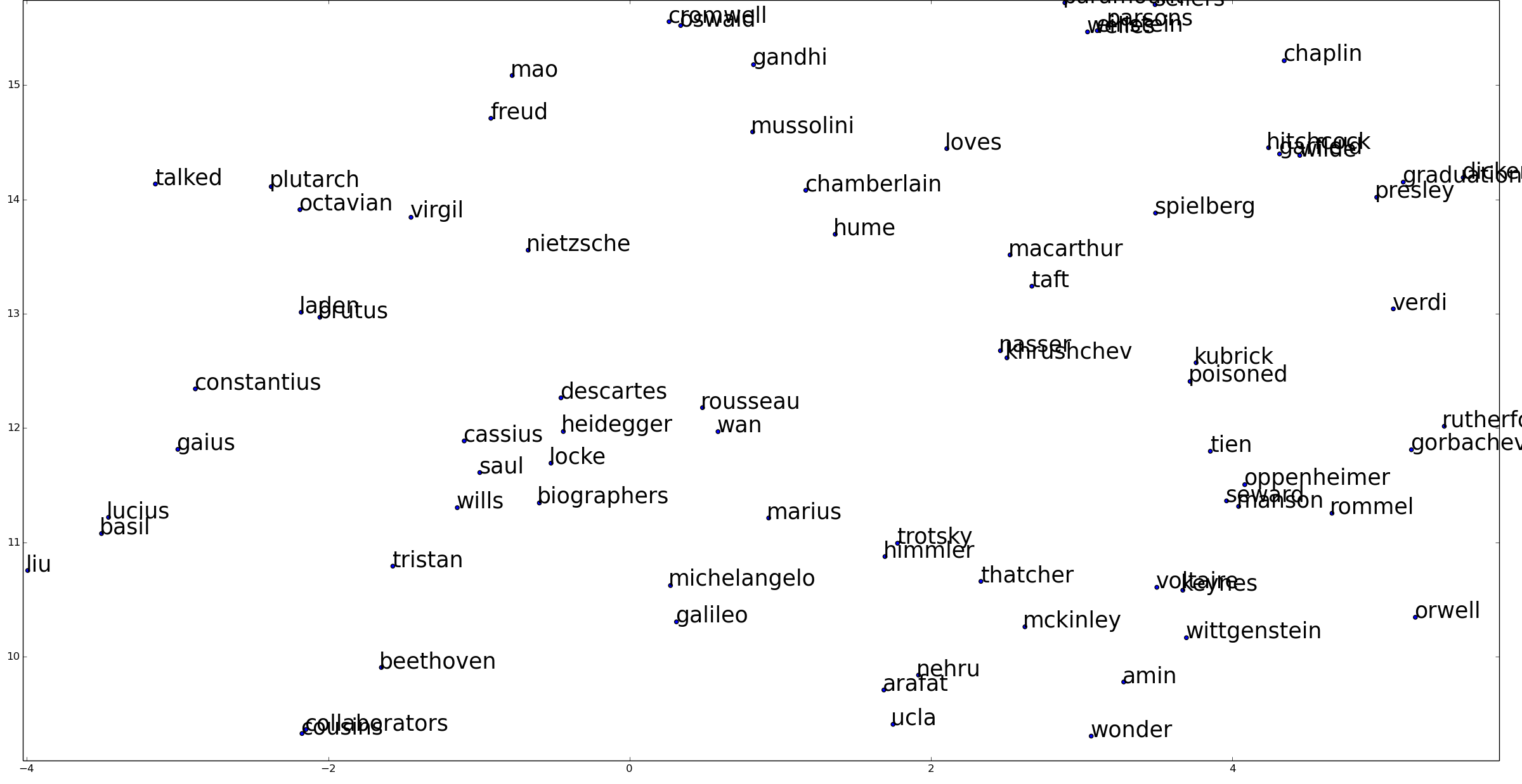}
  \caption{The algorithm has grouped people, suggesting that the algorithm is able to recognize syntactical similarities by looking at the context words. We can also differentiate groups of philosophers, movie directors and famous Roman and Greek people.}
  \label{fig:peopleTSNE}
\end{figure}

In Figure \ref{fig:peopleTSNE} we show an interesting sector extracted from the reduced matrix cartesian plot.

\subsection{The myth behind word analogies and Hash2Vec for word analogies}

Word analogies have been used to show the expressiveness power of a word embedding model \cite{MikolovAnalogReasoningTask}. GloVe and Word2Vec produce embeddings where a linear relationship exists between analog words. The most popular example is the analogy ``prince is to princess like king is to \dots '' where the model should show that prince-princess $\equiv$ king-queen. In \cite{Glove} the authors claim that models that capture this linearity have a superior understanding to others. In contrast, \cite{Levy} show that the word analogy task is just a derivation of similarity. When asking the model ``$x$ is to $y$ like $z$ is to $w$''  we are actually maximizing the dot product of $w(x+y-z)$ which is the same as maximizing $wx + wy -wz$ so the word analogy task is actually asking the model about words that are similar to $x$ and $y$ but different to $z$.  

Word embeddings like GloVe or Word2Vec are not capable of capturing words that are different to a given word so they usually fail to find antonyms, this means that the word analogy task is strongly dependent on the words that are similar to the positive words in the query.

\begin{table}[]
\centering
\caption{Word ``linearity''}
\label{linearity}
\begin{tabular}{@{}llll@{}}
\toprule
x     &  is to y & like z & is to ... \\ \midrule
Paris & France  & Moscow & \textbf{Russia}    \\
Cow   & Milk    & Pig    & \textbf{Meat}      \\
Glass & Glasses & Horse  & \textbf{Horses}    \\
Nice  & Ugly    & Small  & \textbf{Large}     \\ \bottomrule
\end{tabular} 

\end{table}

Table \ref{linearity} shows how Hash2Vec is able to solve some word analogy tasks. The vectors are not trained to preserve linearity in the word analogy task, the original matrix captures some of these properties and therefore Hash2Vec can use them.
The case of ``Cow is to Milk like Pig is to Meat'' is not about the model being ``clever'' or capturing a magic linearity but just a case of ``Meat'' being used in similar contexts as ``Cow'' and ``Milk''. 

What the embedding models really capture is then not exactly a linearity. Good embeddings are capable of capturing different levels of similarity simultaneously. We have shown examples of semantic similarity, but there are other similarities captured by the model like ``diction'' similarity where words that are spelled similarly are close together in the embeddings. 
\section{Conclusions}
In this work we detailed a very simple algorithm that is able to construct word embeddings in linear time. The algorithm does not require training and has minimal memory footprint. We showed the results of the algorithm to be comparable to GloVe \cite{Glove} in the word similarity and analogy tasks, which is one of the state of the art algorithms for word embeddings. The algorithm can also be used to approximate the word co-occurrence matrix to train models like GloVe or EMF with minimal memory consumption.  

The results show that feature hashing is a very powerful technique when applied to text as it can reduce the dimensionality of word vectors from millions to only hundreds while capturing the semantic of each token in the vocabulary. 

Hash2Vec can be used as a very fast way to construct word embeddings with minimal code for streams or dynamic corpora and then use these embeddings in NLP applications.

%
%

\clearpage
\addtocmark[2]{Author Index} 
\renewcommand{\indexname}{Author Index}
\clearpage
\addtocmark[2]{Subject Index} 
\markboth{Subject Index}{Subject Index}
\renewcommand{\indexname}{Subject Index}
\end{document}